# DreamStruct: Understanding Slides and User Interfaces via Synthetic Data Generation

Yi-Hao Peng, Faria Huq, Yue Jiang, Jason Wu, Amanda Xin Yue Li, Jeffrey Bigham, Amy Pavel

Carnegie Mellon University, University of Texas Austin, Aalto University
{yihaop,fhuq,jasonwu2,xal,jbigham}@cs.cmu.edu, yue.jiang@aalto.fi, apavel@cs.utexas.edu

**Abstract.** Enabling machines to understand structured visuals like slides and user interfaces is essential for making them accessible to people with disabilities. However, achieving such understanding computationally has required manual data collection and annotation, which is time-consuming and labor-intensive. To overcome this challenge, we present a method to generate synthetic, structured visuals with target labels using code generation. Our method allows people to create datasets with built-in labels and train models with a small number of human-annotated examples. We demonstrate performance improvements in three tasks for understanding slides and UIs: recognizing visual elements, describing visual content, and classifying visual content types.

## 1 Introduction

Computationally understanding the underlying structures of visual designs, such as presentation slides and user interfaces (UIs), enables machines to interpret and describe the visuals for people who are blind [44, 51, 72, 84], retarget layouts to new devices [37, 38] and personalize content based on user ability [20, 54, 56, 77]. However, building the underlying machine learning models that enable these capabilities requires labor-intensive data collection and annotation, which must be performed for each type of input.

We present a method to generate synthetic, structured visuals by generating and rendering code (Figure 1). Our approach involves three phases: first, we create design ideas with a large language model (LLM) based on the design principles and targeted tasks; second, we generate labeled *declarative language* such as HTML code based on these design ideas to represent structured visuals; third, we filter, post-process, and render the code to produce finalized annotated datasets. While our method is applicable to various types of structured visuals, we apply our method to two application domains that lack high-quality, public datasets for computational modeling: presentation slides and UI screenshots.

For each application domain, we address three visual understanding tasks: *i)* element recognition, *ii)* image captioning, and *iii)* image classification. Compared to traditional data collection and annotation methods, our approach creates synthetically-annotated training data on demand, which is more scalable and can be generalized or fine-tuned for real-world use cases. Compared to automated approaches such as crawling and metadata extraction [78, 81], our approach can apply to more visual domains (*e.g.,* presentation slides) and produce more types of annotations (*e.g.,* semantic descriptions, visual classification).

Using our method, we generate two synthetic datasets: one with 10,053 labeled slides (*DreamSlides*) and another with 9,774 labeled user interfaces (*DreamUI*). We evaluated our method by training machine learning models on our synthetic datasets and comparing the performance with two types of baseline models: models fine-tuned with existing human-annotated data, and non-finetuned models with notable zero-shot performance (e.g., LLaVA [47], Qwen-VL [7]). Our results showed that for element recognition, models trained with synthetic data demonstrated better performance compared to those trained only on human-annotated data (achieved 10.95% improvement by training on synthetic slides, and 5.20% improvement by pretraining on synthetic UIs). For image captioning, our synthetically finetuned models achieved better average win rates against baseline models of 68.9% for slides and 67.1% for UIs. For image classification, our synthetically finetuned models also surpassed their counterparts trained on human-annotated datasets by 18.4% for slide images and 11.3% for UI images. Our results demonstrate the potential of learning structured visual semantics and representations using synthesized code abstraction.

## 2 Related Work

We review literature in three related areas: *i)* understanding slides, *ii)* understanding UIs, and *iii)* synthetic data generation for representation learning.

### 2.1 Understanding Slides

Previous research developed human-annotated datasets and models for interpreting visual and textual content in presentation slides, enabling flexible presentation consumption and authoring experiences [37, 52, 53, 55, 57]. The SpaSe [25] and WiSe [26] datasets were the two early works that provide manually curated object segmentation masks for 2K slides from a large-scale online presentation platform [5]. Further advancements in slide modeling have involved annotating bounding boxes for elements on more than 50K slides, facilitating applications such as visual question answering [67] and sketch-based slide retrieval [35]. In addition to static slide documents, the Lecture Presentation Multimodal Dataset [40] includes annotations for approximately 8.6K graphic elements across 9K slide frames derived from lecture videos. The FitVid dataset [37] extends the data annotations by labeling both individual-level and semantic-group-level elements (e.g., headers, footers, bullet text boxes), totaling 26K element annotations across 5.5K slides. Unlike earlier methods that relied on human annotators, our approach generates synthetic slides with relevant metadata on demand, minimizing the need for manual annotations.

### 2.2 Understanding UIs

The introduction of large-scale UI datasets by prior works [9, 16, 39, 86] enabled the development of data-driven approaches for computational UI modeling – including UI element recognition and semantic grouping [12–14, 50, 82, 86], assessments of interactivity of UI elements [60, 64, 78], and evaluating agentic workflows for UIs [17, 34, 87]. However, collecting such large-scale datasets is often time-consuming and labor-intensive. Unlike these methods that rely on manual collection and human annotation for real UI data, our method uses synthesized UIs as training data, allowing people to construct datasets flexibly while reducing time and manual effort. We assess our method by assessing the models trained with our synthetic data on a couple of existing UI datasets, including the dataset for semantic element and grouping recognition [10], screen captioning [74], and screen design pattern classification [43].

### 2.3 Synthetic Data Generation for Representation Learning

The concept of using synthetic data to learn real-world distributions has been applied across various domains including robotics (sim-to-real) [6, 15, 59] and healthcare [36, 68]. With the advent of generative models, recent research has increasingly used generative models to create synthetic data for visual and language representation learning [8, 19, 32, 33, 69, 73, 75]. One particularly relevant work is WebSight [29], which presents a synthetic dataset of approximately 820K `HTML` code examples generated from screen descriptions. Similar methods have been employed to generate synthetic data for other modalities, such as vector graphics [58, 61], charts [24, 70], animation [71], and 3D models [18, 83]. While these papers generate code to synthesize visuals, their objectives are fundamentally different from ours. They focus on translating between pixels and code abstractions (pix2code and code2pix) without explicitly generating task-specific underlying structures, which limits their applicability in understanding structured visuals (slides and UIs). In contrast, our method generates task-specific metadata alongside code, providing a fine-grained training source for structured visual representation learning.

## 3 Methods

Instead of relying on traditional methods for building visual understanding datasets, which require human annotation of inputs (*e.g.,* screenshots), DreamStruct begins with an abstract specification of the desired annotations and constraints, then generates the corresponding visual inputs as renderable code. In this section, we describe our methodology for creating synthetically labeled slides and UIs. We first explain our generation principles, followed by the generation pipeline and the analysis of our synthetic datasets.

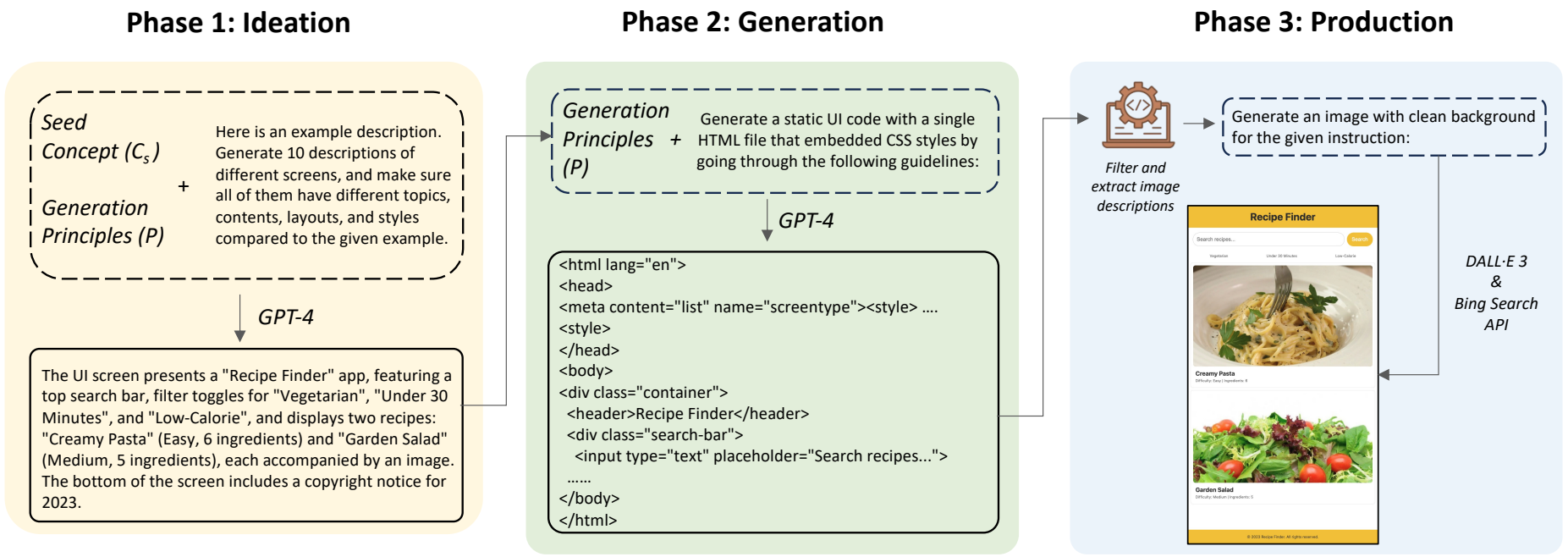

**Fig. 1:** Overview of the DreamStruct pipeline. From left to right: In the *Ideation Phase*, screen descriptions for slides and mobile UIs are generated. In the *Generation Phase*, corresponding code is produced based on these descriptions, guided by the design principles and targeted labels. Finally, in the *Production Phase*, visual placeholders are replaced with actual data to render the final slides and UIs.

### 3.1 Generation Principles

To ensure the generated structured visuals can help model the data in-the-wild, we defined *design principles* and *goals* for our generation process. Establishing a set of rules helps us formulate better prompt instruction [66] that embodies the desired outcomes. We specifically propose the following principles ($P$):

*P1.* **Consistency with existing visual design guidelines:** The generated design and layout should abide by the established design guidelines followed by practitioners. For example, UI buttons should be of dimension $\geq 44 \times 44$ pt [4], and the font size of the main text content on the slides should be bigger than 24 pt [48].

*P2.* **Consistency with existing visual design patterns:** The generated visual design should reflect common design patterns ( extite.g., visual design themes or layouts). For example, the login and setting screens are common classes for mobile UI design [43, 62], and the title slides and section slides are common patterns for presentation design [23, 27, 63].

*P3.* **Consistency with target data labels:** The generated metadata should align and remain consistent with all specified data labels for slides and UI benchmarks. The principle applies to both element-wise [9, 37] and screen-wise semantics [25, 37, 43]. For instance, an interactive swipe feature in UIs should be categorized as 'pageindicator'.

### 3.2 Generation Pipeline

Our generation pipeline consists of three steps (Figure 1): (1) *Ideation*: Create design concepts for slides and UI. (2) *Generation*: Generate code with target labels using design concept descriptions. (3) *Production*: Filter, post-process, and render the code to produce the complete slides and UIs.

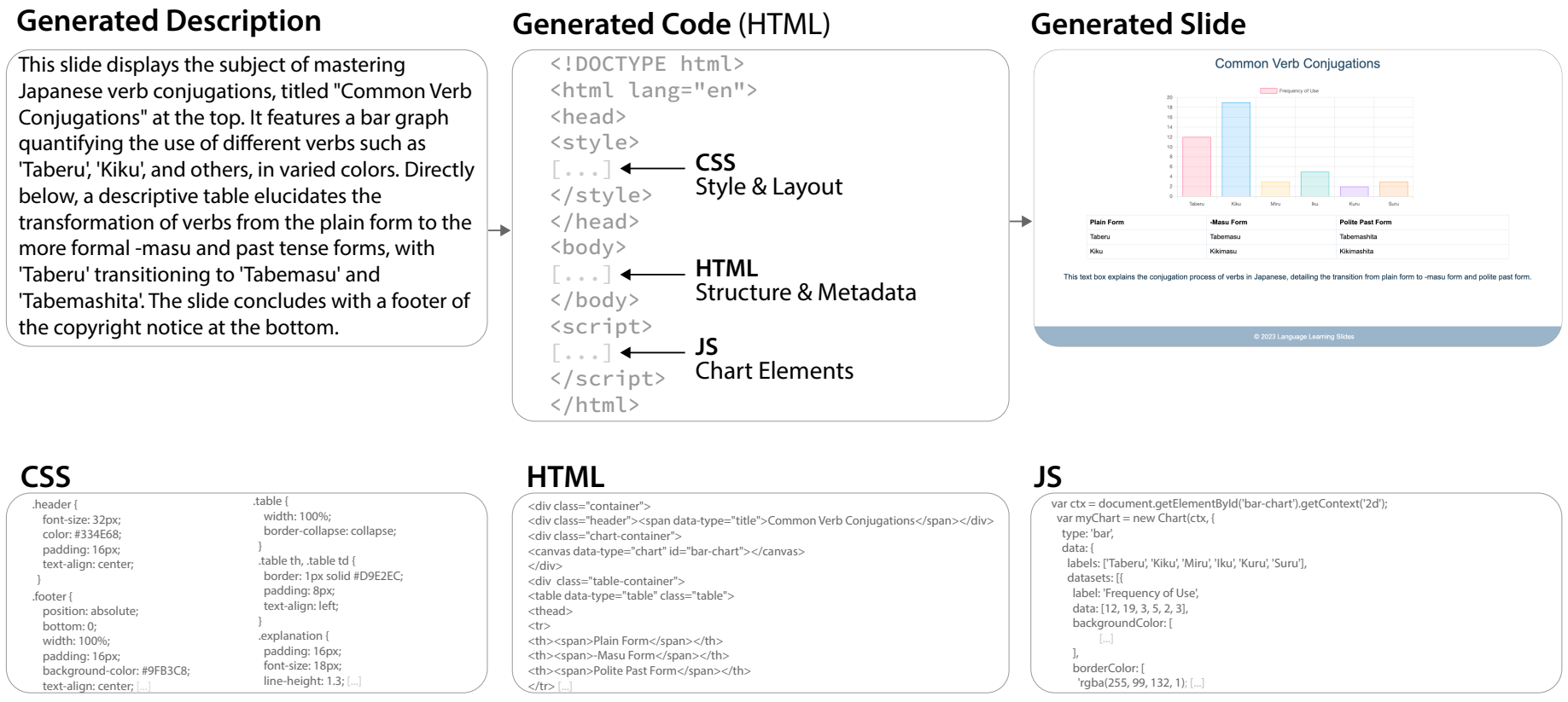

**Fig. 2:** An example of a generated slide description with the corresponding generated code and generated slide (top) with example `CSS, HTML, and JavaScript` excerpts from the generated code (bottom).

**Phase 1: Ideation.** We generate a corpus of design concepts ($C_d$) for slides and UIs by prompting an LLM with our generation principles ($P$) and a set of seed design concept examples ($C_s$). To obtain our seed design concept examples ($C_s$), we examine real-world instances of slides and UIs [37, 40, 43]. From the examination, we select 20 diverse samples for both slides and UIs. We then manually write descriptions for all the samples based on prior visual description guidelines [22, 57]. Each human description provides a summary of the visual content, style, and layout (i.e. a design concept). For example, *The slide titled "By the Numbers" presents three key statistics on disability representation, each in a separate column with a corresponding icon. The statistics cover general disability prevalence, volunteer participation, and board representation. A source reference is provided at the bottom. The slide uses a dark grey color scheme.* We then prompt an LLM with $P$ (our design guidelines, design patterns, and target labels) as our instructions and $C_s$ as our few-shot examples to ensure that the generated corpus of design concepts $C_d$ are consistent with $P$ [49] and the style of human-written descriptions. In other words, $\forall c_d \in C_d,\ c_d \sim \mathcal{G}_{t=0.5}(C_s, P)$, where $\mathcal{G}$ denotes the LLM (`gpt-4-1106-preview` in this case) and $t$ denotes the temperature parameter used in $\mathcal{G}$ during decoding. In this manner, we achieve a high-quality and diverse corpus $C_d$ where the descriptions maintain the pairwise BERTscore [85] $< 0.7$, a threshold that was used to determine the diversity of model-generated descriptions [73].

**Phase 2: Generation.** We approach visual layout creation for slides and UIs as a code generation task. We use `HTML` for layout creation because of its semantic structure, which allows label embedding during generation. Unlike text-to-image models that generate visual data directly from descriptions, our code-based method offers flexible visual generation and detailed annota-tions. To ensure the generated code for $\forall c_d \in C_d$ has the desired annotation formats and adheres to design guidelines, we incorporate the principles, $P = \{P1, P2, P3\}$ in the input instructions (see supplementary materials for the complete prompts). Specifically, we instruct the LLM to add additional semantic HTML attributes as the targeted annotations. For instance, in the element recognition task, individual element types are labeled with `data-type` attributes (*e.g.,* `<span data-type="text">this is a text element</span>`). For image classification, we label the screen type as attributes of global `<meta>` tags (*e.g.,* `<meta content="login" name="screentype">`). For graphical elements that can be produced via image retrieval or generation (e.g., images, icons, and diagrams for our selected domains), we ask the model to generate an image element with a text description of the intended image content as alternative text (`<alt="Fresh salad bowl">`), a placeholder image source (`<img src="placehold.co">`), and `width` and `height` values. For more complex graphical elements that are challenging to create with image retrieval or text-to-image models, we instruct the LLM to use external code libraries to create the visuals (e.g., charts in our selected domains and tasks, we selectively use `chart.js` [3], `chartist.js` [2], and `chart.css` [1] as our generation libraries). At this step, we have an updated corpus of pairwise descriptions and annotated code that both follow our principles, $C_h = \{\mathcal{G}_{t=0.3}(c_d; P),\ \forall c_d \in C_d\}$.

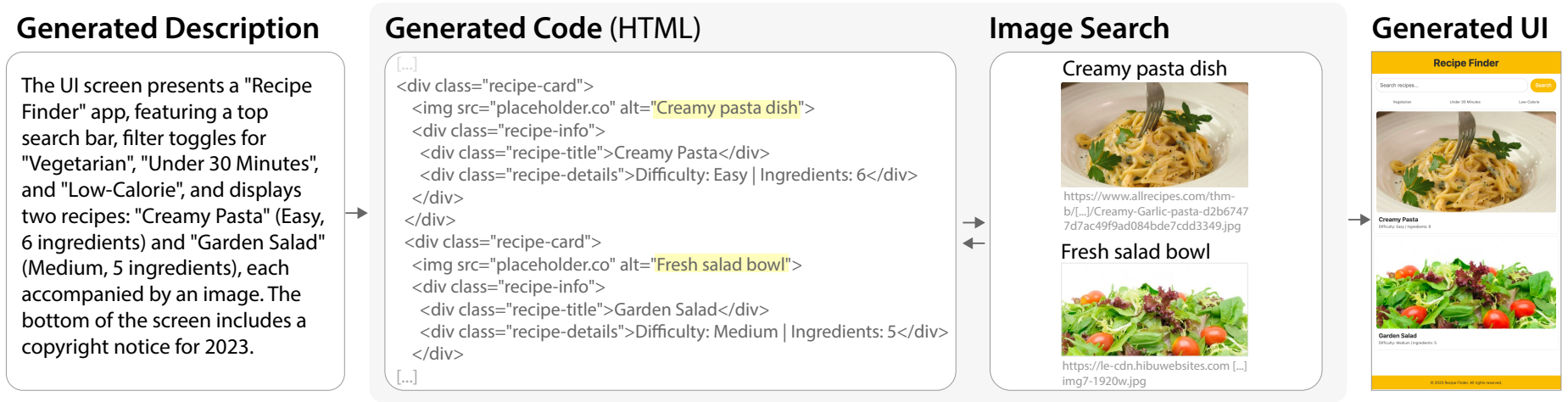

**Fig. 3:** An example of a generated UI description with the corresponding generated code and generated UI. In the *Production* phase, our pipeline extracts alternative text and placeholder image links and uses Bing Search API or `DALL·E` to get a link or image to replace the placeholder.

**Phase 3: Production.** The final step involves post-processing, filtering, and rendering the generated code for slides and UIs. We use the `<alt>` value of each graphical element placeholder as the basis for visual updates. For simple graphical elements (*e.g.,* images and icons in our selected domain), we query the Bing Search API to retrieve image sources with transparent backgrounds. For complex graphical elements (e.g., diagrams in our domain chosen), we prompt text-to-image model DALL·E to generate image sources. We then download each image and replace the placeholder source with a link to the new image source. Besides graphical elements rendering, we take additional heuristic-based post-processing steps to improve the quality of generation: 1) removing background fill colors when a background image is present, 2) adding width and height for <img> elements (*e.g.,* replacing `<img src="https://via.placeholder.com/280x400">` with `<img style="height: 400px; width: 280px">`) for proper fitting, and 3) keeping sliding menus open on page load to ensure visibility. To reduce the low-quality samples caused by major element occlusion or invalid `<img>` references, we compute the CLIP score between the rendered screens and their design concepts, and we filter out examples with a CLIP score below 0.3. We choose the threshold based on the average image-text alignment for LLM-based text-to-code generation [80]. In total, we filtered out 215 (2.1%) slide and 336 (3.36%) UI samples. The final `HTML` is rendered within viewports of 1280 × 720 for slides and 628 × 1118 pixels for UIs. Our updated corpus can be represented as $C_{dhm} = \{\mathcal{R}(\mathcal{F}(c_h)) \mid \forall c_h \in C_h \text{ and } \forall c_d \in C_d, \text{CLIP}(\mathcal{R}(\mathcal{F}(c_h)), c_d) > 0.3\}$, where $\mathcal{R}$ and $\mathcal{F}$ are the rendering and filtering functions.

| | **Slides** | **UIs** |
|---|---|---|
| # of Descriptions | 10,268 | 10,000 |
| # of Generated Code (after filtering) | 10,053 | 9,774 |
| # of Generated Images and Icons | 15,593 | 30,518 |
| # of Generated Charts | 2128 | - |
| # of Avg. Element | 6.46 | 11.08 |

**Table 1:** Statistics of DreamStruct dataset.

### 3.3 Synthetic Dataset

Table 1 presents the statistics for the data generated in DreamStruct. We created a total of 10,053 slide-code pairs (DreamSlides) and 9,774 UI-code pairs (DreamUI). We quantitatively compare DreamStruct with human-annotated datasets in terms of descriptions and element compositions. DreamStruct provides significantly more detailed descriptions for slides and UIs, averaging 57.23 words and 19.66 named entities per description, compared to 5.74 words and 3.37 named entities in human-annotated datasets. The differences in element compositions are marginal, including the number of images per sample (DreamStruct=2.32, human=2.48), charts per sample (DreamStruct=0.21, human=0.07), diagrams per sample (DreamStruct=0.24, human=0.12), and total elements per samples (DreamStruct=8.74, human=8.83). Figure 4 illustrates the distribution of elements in our synthetic slide and UI datasets compared to ground-truth datasets. Even without explicitly controlling the distribution of element classes during generation, the element distribution in our synthetic data aligns with real-world datasets [9,37]. An exception is the 'upper task bar' in our synthetic UI dataset, which displays the time and internet signals and is not typically part of the UI content. Consequently, the upper task bar appears infrequently in our dataset, as shown in Figure 4b.

`https://github.com/yihaop/dreamstruct`

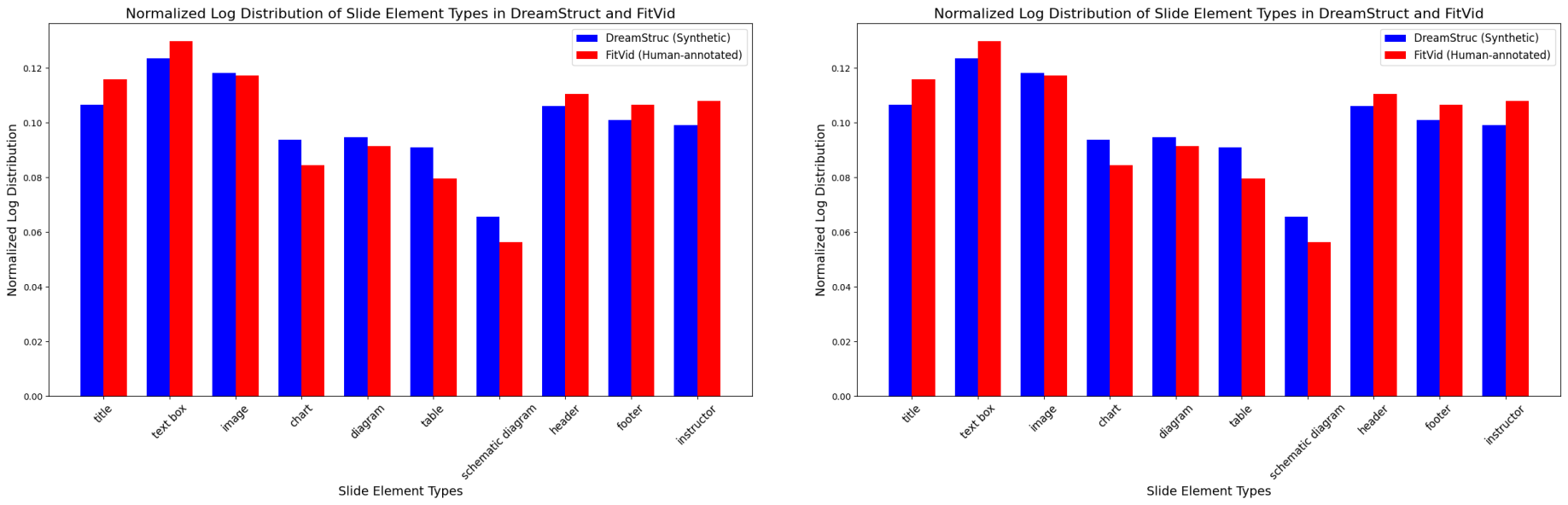

**(a)** Normalized Log Distribution of Slide Element Types in DreamStruct and FitVid

**(b)** Normalized Log Distribution of UI Element Types in DreamStruct and VINS

**Fig. 4:** Normalized Log Distribution of Slide and UI Elements

# 4 Experiments

We evaluate our synthetic method by training and pretraining models on our generated datasets for slides and UIs and comparing to strong baselines consisting of models trained on human-annotated data and publicly-available models. The chosen models are designed to perform element recognition, image captioning, and image classification on screenshots, which are essential tasks for the computational understanding of slides and UIs.

## 4.1 Element Recognition

Element recognition identifies and outlines elements in slides and UIs from pixel input. The recognition results can be applied to domains such as accessibility [86], design [9], and device retargeting [37]. For our base object recognition models, we selected several vision backbones, including CenterNet2 [88], Deformable-DETR [89], and EfficientDet-D0 [65].

**Slide Element Recognition.** For slide element recognition, we used a lecture design dataset [37] as our benchmark, with a split of 70%, 15%, and 15% for training, validation, and testing respectively. We ensured that slides from the same presentation videos were grouped in the same split. The separation is to prevent the models from training and testing on similar data as slide frames from the same video are typically similar to each other. The dataset consists of 5.5K slides and includes 12 types of semantic elements, across 14 domains (ranging from engineering to social science). We merge classes that are semantically close to each other (*e.g.,* 'figures' and 'pictures' into 'images', 'handwritten' into 'text box'), resulting in 10 types of element types: 'title', 'text box', 'image', 'chart', 'diagram', 'table', 'schematic diagram', 'header', 'footer', and 'instructor'. The dataset provides bounding boxes for both individual elements (titles, text boxes, images, charts, diagrams, tables, schematic diagrams, instructors) and container-type elements (headers and footers), and introduces unique labels such as "instructor" that are not commonly found in other related datasets [25, 26, 35, 40].

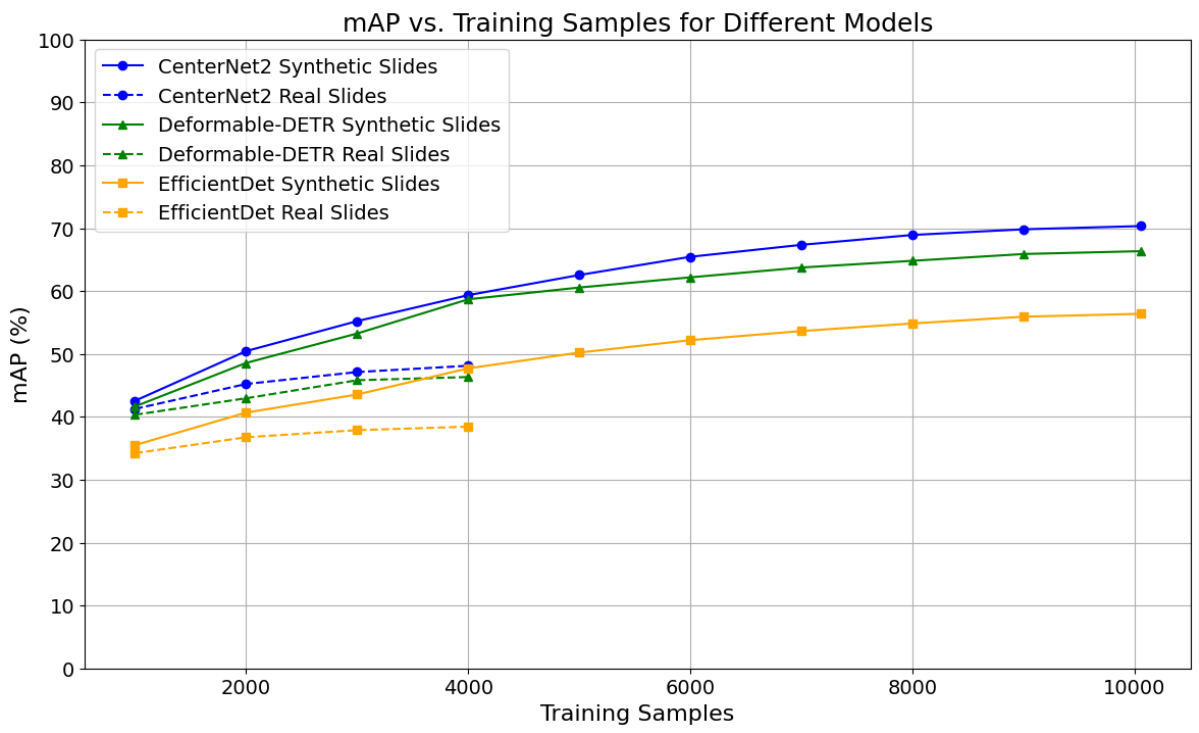

**Fig. 5:** Performance of slide element recognition when adding training samples

Our findings show that models trained on synthetic slides achieved higher mean Average Precision (mAP) than those trained on human-annotated data. Specifically, models trained on synthetic data achieved an average of 55.3% mAP compared to 44.3% mAP for models trained on an equivalent sample size of human-annotated data (full training set). The limited performance of models trained on human-annotated data was due to a lack of visual design variation, as many training samples were derived from the same video. In contrast, our synthetic data generation method ensures a diverse dataset, resulting in a more generalized data distribution. Figure 5 shows the performance curve as we gradually added more synthetic training samples. Performance gains plateaued and peaked at an average of 64.4% mAP when training model with a synthetic dataset twice the size of the original human-annotated set. Our performance analysis of individual element types revealed that our models excelled in recognizing tables and charts due to the diverse data generated through code variations coupled with image retrieval and generation.

**UI Element Recognition.** For UI element recognition, we select the mobile UI design dataset [10] as our benchmark, with the train, val, and test splits of 70%, 15%, 15% derived from prior work [81]. The dataset contains 5K mobile UI screens along with 12 classes of elements (*e.g.,* 'text', 'image', 'text button', 'icon', 'input field', 'switch', 'checked view', 'background image', 'sliding menu', 'upper taskbar', 'page indicator', 'popup window') across the screenshots of iOS and Android applications as well as UI mockups. The dataset provides bounding boxes for both individual elements (text, images, text buttons, icons, switches, checked views, background images, page indicators) and container-type elements (input fields, upper task bars, sliding menus, popup windows).

To address the element distribution gap between the synthetic and the real-world data, we adopted a two-stage training approach. Initially, we used our synthetic dataset to pretrain the models, which allowed us to establish a foundational set of weights. This pretraining step is crucial as it initializes the model's weights with knowledge gained from the synthetic UI data, despite its distributional differences from real-world data. After pretraining, we then finetuned the models on human-annotated data. This finetuning phase adjusts the pretrained weights to align more closely with the real-world data distribution, thereby enhancing the model's performance on tasks that reflect real-world scenarios. Our findings show that using synthetic UIs for pretraining, followed by finetuning with only 50 samples of human-annotated data, resulted in an average mAP improvement of 5.2% compared to training solely with the same set of human-annotated samples (from an average mAP of 47.2% to 52.4%). As we gradually added human-annotated samples, the pretraining step continued to show benefits, with an average mAP improvement from 78.8% to 81.8% when finetuning on the full human-annotated training set (Figure 6).

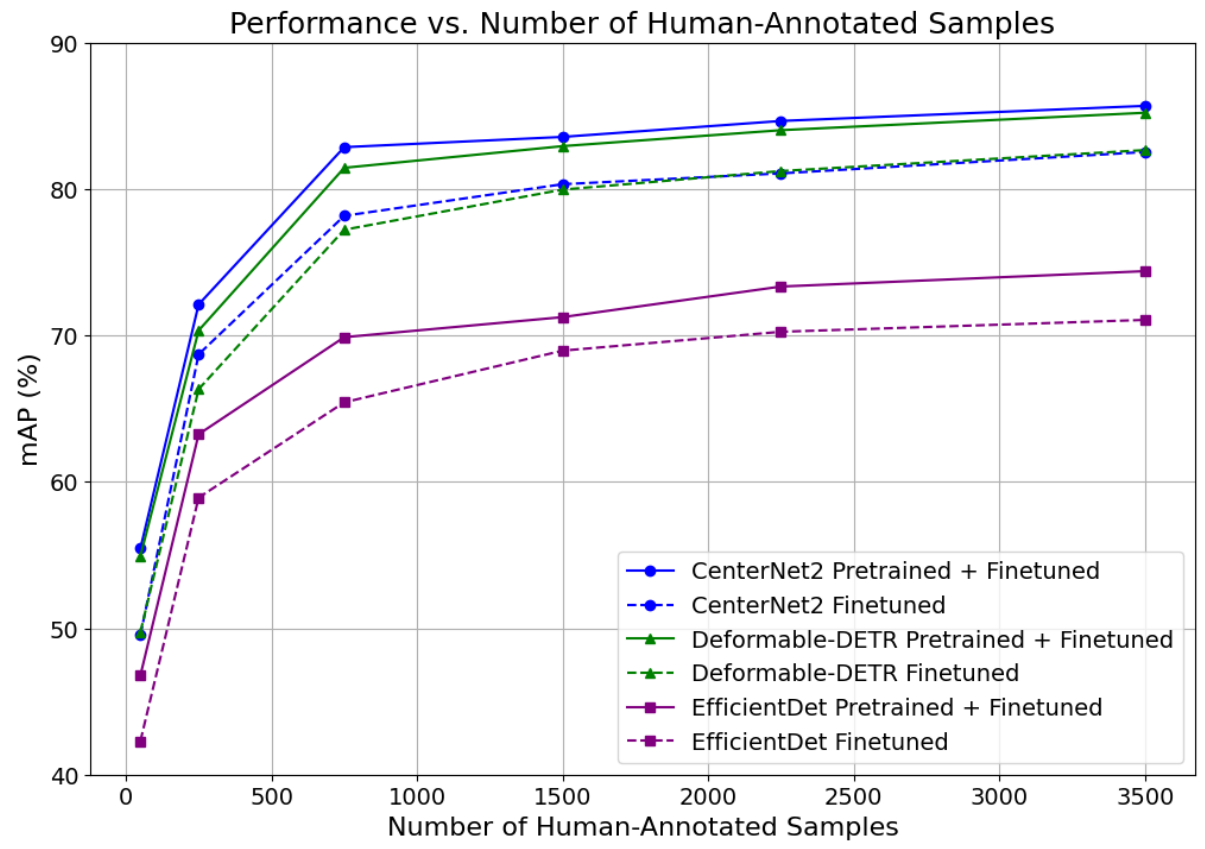

**Fig. 6:** Performance of UI element recognition when adding training samples

## 4.2 Image Captioning

Image captioning refers to the task of generating descriptions for the screenshots of slides and UIs. The descriptions can support blind and low-vision individuals in accessing information from visually structured images [41, 57, 74]. We evaluate the quality of descriptions generated from four vision-language model (VLM) variations: our model that fine-tunes LLaVA-1.5-13B using the synthetic datasets; the original LLaVA model without fine-tuning; the vision language models including LLaVA or Pix2Struct fine-tuned with existing human-written captions [35, 74]; and GPT-4V (zero-shot). We conducted manual pairwise comparisons with 6 human annotators. Specifically, we first asked the annotators to label the same 50 slides and UI samples and we ensured the consistency of the preference ratings across human annotators based on the prior works on image captioning ratings [21, 22]. We then let each human rater rate 2K comparisons randomly sampled from two unique model outputs, resulting total of 10,392 comparisons and preference selections.

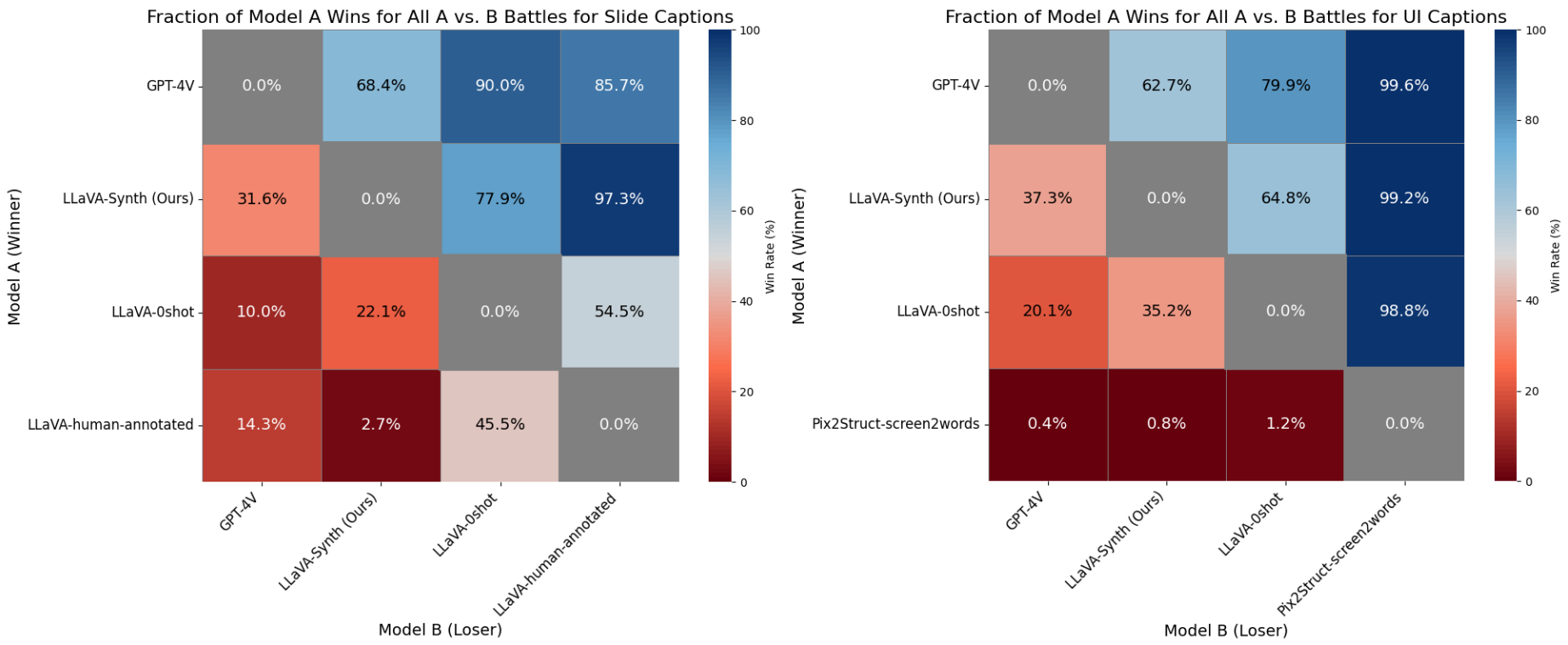

**(a)** Win rate matrix of slides

**(b)** Win rate matrix of UIs

**Fig. 7:** The matrices showing each pair-wise win rate comparison across models

**Slides Image Captioning.** For slide description, We randomly sampled 732 slides from existing lecture presentation slide datasets [37, 40] as our test set. Figure 7a presents the pairwise win rates between our methods and baseline conditions including LLaVA zero-shot, LLaVA fine-tuned with existing slide descriptions [35], and GPT-4V. The model fine-tuned with our synthetic image-description pairs surpasses those fine-tuned with current description datasets (77.9% win rate) and those derived from zero-shot inference (97.3% win rates). Our model, though smaller, still achieves a 31.6% win rate compared to GPT-4V.

**UI Image Captioning.** For UI description, We randomly sampled 1000 UIs from existing UI datasets [10] as our test set. Figure 7b presents the pairwise comparison results between our methods and baseline conditions (including LLaVA zero-shot, the vision-language model Pix2Struct [42] finetuned on existing screen2word dataset [74], and GPT-4V). The model fine-tuned with our synthetic image-description pairs surpasses those fine-tuned with existing description datasets (64.8% win rate) and those generated from zero-shot inference (99.2% win rate). Compared to GPT-4V, our model achieves 37.3% of the win rate despite our base model size being smaller than GPT4-V.

### 4.3 Image Classification

Image classification involves identifying the themes and designs of individual slides and UIs across subjects or visual patterns. By categorizing the themes, the machine can perform tasks like detecting lecture topics or chapters [11, 46, 53], and offering UI design recommendations [10, 43]. We chose two VLMs, LLaVA-1.5-13B [47] and Qwen-VL-Chat-7B [7], as our base classification models. We compared the classification performance across multiple conditions: our models that finetuned based models using synthetic datasets; the original base models (zero-shot); the based models finetuned with existing human-annotated data; and two off-the-shelf VLMs: GPT-4V and Gemini-1.0-Pro. For the human-annotated dataset, we maintained an 80% training and 20% testing split for benchmarking. Notably, the current image classification datasets for either UIs or slides only comprise around 1,000 samples or fewer.

| **Model** | **Slides** | **UIs** |
|---|---|---|
| GPT-4V | 0.772 | 0.598 |
| Gemini-1.0-Pro-Vision | 0.700 | 0.533 |
| LLaVA synth (ours) | 0.686 | 0.545 |
| Qwen-VL synth (ours) | 0.672 | 0.558 |
| LLaVA zero-shot | 0.462 | 0.395 |
| Qwen-VL zero-shot | 0.473 | 0.382 |
| LLaVA human-annotated | 0.335 | 0.316 |
| Qwen-VL human-annotated | 0.341 | 0.303 |

**Table 2:** The classification accuracy of each model for both slides and UIs (training samples: synth≏5K; human-annotated≏0.5K)

**Slide Classification.** For slide classification, we selected the top 10 topics from previous datasets [37, 40] and combined similar themes, resulting in six topics: psychology, communication, law, public health, computer science, and language learning. We initially sampled 960 slides, excluding irrelevant ones such as 'Thank You' slides, which led to a dataset of 732 slides as our ground truth set. From this human-annotated set, we randomly chose 25 slides per topic, creating a test sample of 150 slides. Table 2 displays the slide classification accuracy. Models trained on our synthetic slides achieved an average classification accuracy of 52.2%, surpassing the original base model (46.8%) and the base models fine-tuned with human-annotated data (33.8%). By expanding the training dataset to 5,033 synthetic slides, our method reached an average accuracy of 67.9%. However, further increases in sample size resulted in diminishing returns (Figure 8a). Despite this improvement, the optimal performance with the current synthetic sample sizes still lags behind GPT-4V by 9.3%.

**UI Classification.** For UI classification, we chose the top-10 design topics from the previous mobile UI design dataset [43]. We merging the class 'News' to 'Gallery' due to its lack of visual correlation with the class name. This yielded categories including 'List', 'Login', 'Settings', 'Menu', 'Media Player', 'Form', 'Profile', 'Tutorial' and 'Gallery', resulting in a dataset of 1020 unique UIs. For evaluation, we randomly selected 25 UIs per category from the dataset, forming a test set of 225 UIs. With a similar number of training samples, our synthetically-trained models demonstrated a 42.3% classification accuracy, outperforming the original base models (38.9%) and base models fine-tuned with human-annotated

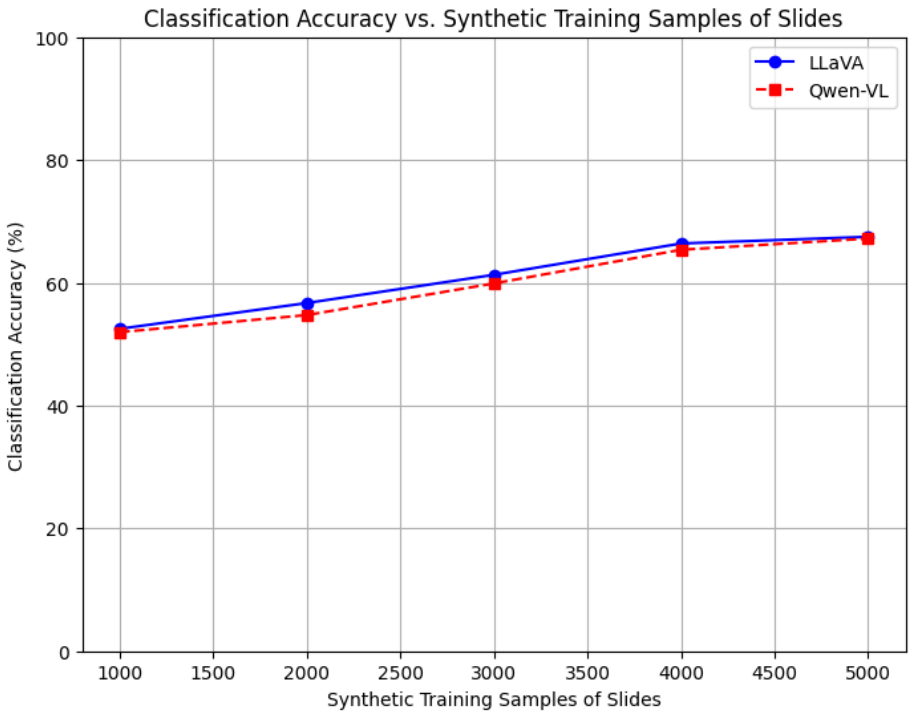

**(a)** Performance of Slide Classification

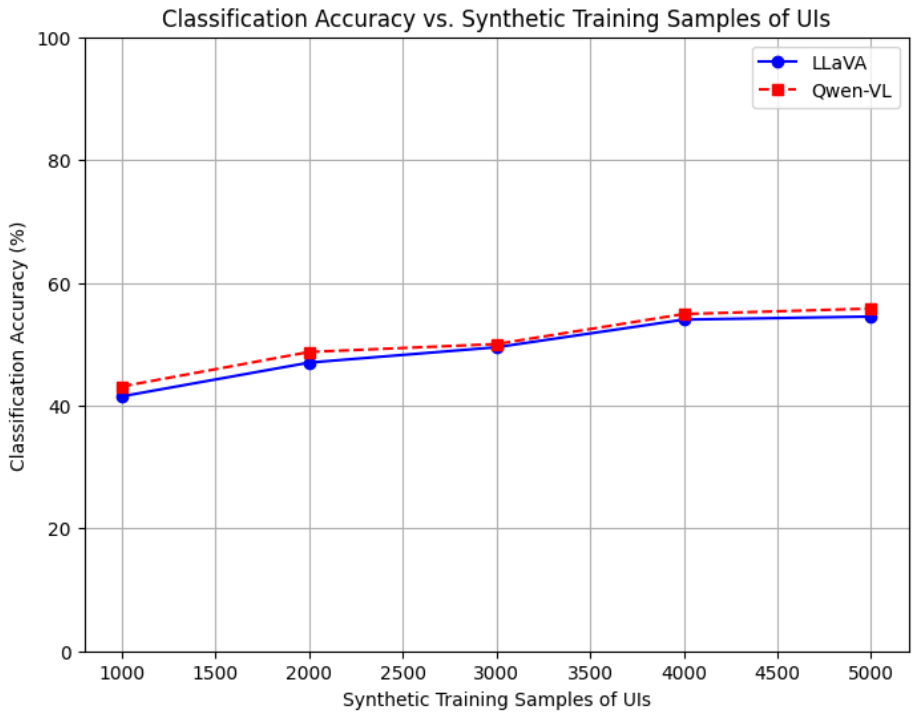

**(b)** Performance of UI Classification

**Fig. 8:** Performance of Slide and UI Image Classification when scaling up synthetic training samples

data (31.0%). With around a ten-fold increase in training samples to 5358 synthetic UIs, the accuracy rose to 55.2%, although further gains plateaued with additional training samples (Figure 8b). The current synthetic training sample sizes result in performance that is 4.7% below that of GPT4-V.

For both slides and UI classification, fine-tuning with human-annotated samples led to performance drops compared to zero-shot inference. This is likely due to the small and class-imbalanced dataset for structured visual classification, limiting the fine-tuned classification performance, as shown in prior work [45, 81].

### 4.4 Summary

The results of our experiments reveal several key findings. First, in domains like slide understanding, where high-quality data is the primary limiting factor, DreamStruct achieves significant performance gains (+20% mAP). In cases like UI understanding, which benefit from larger quantities of human-annotated data, incorporating DreamStruct's synthetic data still leads to substantial improvements in a fine-tuning setting (+5% mAP). Across all tasks and domains, training DreamStruct's synthetic data leads to performance gains and allows open-source models (*e.g.,* LLaVA, Qwen-VL) to approach the performance of larger proprietary baselines (*e.g.,* GPT, Gemini). Our results demonstrate the potential of learning various granularity levels of visual representations through generative code-oriented semantics.

## 5 Limitations

While our synthetic method leads to better performance on couple of downstream tasks for slides and UI understanding, there are still potential concerns and limitations regarding the current data collection scheme and capacities.

**Quality of Code Generation.** Although we specify guidelines for element positioning and spacing (§3.2) during our generation, the generated layouts sometimes fail to follow the specific guidelines, or the generated code adds the metadata into the wrong place. This results in `CSS` styling errors as well as overflow and overlaps in the generated layouts. For example, the data-type property was mentioned inside `CSS` tag 1,183 times across 354 screens, although it is not a valid `CSS` property. While we could fix these minor errors by parsing the `HTML` during post-processing (§3.2), future work should be mindful that such errors can occur in synthetic data generation framework [66]. To improve the code generation quality, future work can incorporate additional domain-specific filtering models [79, 80] as reward functions to enhance the code-based visual generation results.

**Limitation of Visual Content Generation.** Currently, we use the image alternative text to generate the visual content using Bing Image Search and `DALL·E` (§3.2). Since we do not pass any additional contextual information for the screen, the generated images may not be consistent in style throughout the layout. Future work can extend the visual content generation process by including additional context [28] and visual asset libraries.

**Potential Negative Impact of the Work.** While our work makes it easier to improve understanding of structured visual content like slides and UIs, there might be the potential risk of misusing the DreamStruct pipeline for false and misleading content creation [76]. We do not endorse the use of DreamStruct to create manipulative content. Future work can investigate the potential risk of synthetic data generation and design a more robust generation pipeline that handles copyright issues. Our work may be used for accessibility purposes (especially for the *accessible creativity* domains [30, 31, 55, 57]). For example, blind people may use element recognition to traverse a slide screenshot from a lecture video using their screen reader [52] and may not be able to detect when our model produces errors (*e.g.*, misses an element). Future deployments should warn users of potential errors when using such models, and future work should continue to improve the accuracy of such models.

## 6 Conclusion

We present a method that uses code generation to create synthetic, structured visuals with embedded target labels. Our approach offers an efficient alternative to traditional, labor-intensive data collection and annotation methods for understanding presentation slides and user interfaces. By generating and utilizing synthetic datasets, we have achieved notable improvements in machine learning models' performance across various tasks, including element recognition, image captioning, and image classification. Our findings highlight the potential of synthetic data to enhance computational models' capabilities in interpreting structured visuals and building the next generation of intelligent interfaces.

## Acknowledgements

We thank the reviewers for their feedback and Yuyu Lin for her major support in our study. This work is funded in part by the National Science Foundation.

## References

1. chartcss. https://chartscss.org/, accessed: 2024-03-07 6
2. chartistjs. https://gionkunz.github.io/chartist-js/, accessed: 2024-03-07 6
3. chartjs. https://www.chartjs.org/, accessed: 2024-03-07 6
4. Human interface guidelines. https://developer.apple.com/design/human-interface-guidelines/ (2024), accessed: 2024-03-07 4
5. Araujo, A., Chaves, J., Lakshman, H., Angst, R., Girod, B.: Large-scale query-by-image video retrieval using bloom filters. arXiv preprint arXiv:1604.07939 (2016) 2
6. Azizi, S., Kornblith, S., Saharia, C., Norouzi, M., Fleet, D.J.: Synthetic data from diffusion models improves imagenet classification. arXiv preprint arXiv:2304.08466 (2023) 3
7. Bai, J., Bai, S., Yang, S., Wang, S., Tan, S., Wang, P., Lin, J., Zhou, C., Zhou, J.: Qwen-vl: A versatile vision-language model for understanding, localization, text reading, and beyond (2023) 2, 11
8. Borisov, V., Seßler, K., Leemann, T., Pawelczyk, M., Kasneci, G.: Language models are realistic tabular data generators. arXiv preprint arXiv:2210.06280 (2022) 3
9. Bunian, S., Li, K., Jemmali, C., Harteveld, C., Fu, Y., Seif El-Nasr, M.S.: Vins: Visual search for mobile user interface design. In: Proceedings of the 2021 CHI Conference on Human Factors in Computing Systems. CHI '21, Association for Computing Machinery, New York, NY, USA (2021). https://doi.org/10.1145/3411764.3445762, https://doi.org/10.1145/3411764.3445762 3, 4, 7, 8
10. Bunian, S., Li, K., Jemmali, C., Harteveld, C., Fu, Y., Seif El-Nasr, M.S.: Vins: Visual search for mobile user interface design. In: Proceedings of the 2021 CHI Conference on Human Factors in Computing Systems. pp. 1–14 (2021) 3, 9, 11
11. Che, X., Yang, H., Meinel, C.: Lecture video segmentation by automatically analyzing the synchronized slides. In: Proceedings of the 21st ACM international conference on Multimedia. pp. 345–348 (2013) 11
12. Chen, C., Su, T., Meng, G., Xing, Z., Liu, Y.: From ui design image to gui skeleton: A neural machine translator to bootstrap mobile gui implementation. In: Proceedings of the 40th International Conference on Software Engineering. p. 665–676. ICSE '18, Association for Computing Machinery, New York, NY, USA (2018). https://doi.org/10.1145/3180155.3180240, https://doi.org/10.1145/3180155.3180240 3
13. Chen, J., Xie, M., Xing, Z., Chen, C., Xu, X., Zhu, L., Li, G.: Object detection for graphical user interface: Old fashioned or deep learning or a combination? In: Proceedings of the 28th ACM Joint Meeting on European Software Engineering Conference and Symposium on the Foundations of Software Engineering. p. 1202–1214. ESEC/FSE 2020, Association for Computing Machinery, New York, NY, USA (2020). https://doi.org/10.1145/3368089.3409691, https://doi.org/10.1145/3368089.3409691 3

14. Chen, S., Fan, L., Su, T., Ma, L., Liu, Y., Xu, L.: Automated cross-platform gui code generation for mobile apps. In: 2019 IEEE 1st International Workshop on Artificial Intelligence for Mobile (AI4Mobile). pp. 13–16 (2019). https://doi.org/10.1109/AI4Mobile.2019.8672718 3
15. Chen, Y., Li, W., Chen, X., Gool, L.V.: Learning semantic segmentation from synthetic data: A geometrically guided input-output adaptation approach. In: Proceedings of the IEEE/CVF conference on computer vision and pattern recognition. pp. 1841–1850 (2019) 3
16. Deka, B., Huang, Z., Franzen, C., Hibschman, J., Afergan, D., Li, Y., Nichols, J., Kumar, R.: Rico: A mobile app dataset for building data-driven design applications. In: Proceedings of the 30th annual ACM symposium on user interface software and technology. pp. 845–854 (2017) 3
17. Deng, X., Gu, Y., Zheng, B., Chen, S., Stevens, S., Wang, B., Sun, H., Su, Y.: Mind2web: Towards a generalist agent for the web. Advances in Neural Information Processing Systems **36** (2024) 3
18. Feng, W., Zhu, W., Fu, T.j., Jampani, V., Akula, A., He, X., Basu, S., Wang, X.E., Wang, W.Y.: Layoutgpt: Compositional visual planning and generation with large language models. Advances in Neural Information Processing Systems **36** (2024) 3
19. Fu, S., Tamir, N., Sundaram, S., Chai, L., Zhang, R., Dekel, T., Isola, P.: Dreamsim: Learning new dimensions of human visual similarity using synthetic data (2023) 3
20. Gajos, K., Weld, D.S.: Supple: automatically generating user interfaces. In: Proceedings of the 9th international conference on Intelligent user interfaces. pp. 93–100 (2004) 1
21. Gleason, C., Pavel, A., Gururaj, H., Kitani, K., Bigham, J.: Making gifs accessible. In: Proceedings of the 22nd International ACM SIGACCESS Conference on Computers and Accessibility. pp. 1–10 (2020) 10
22. Gleason, C., Pavel, A., McCamey, E., Low, C., Carrington, P., Kitani, K.M., Bigham, J.P.: Twitter a11y: A browser extension to make twitter images accessible. In: Proceedings of the 2020 chi conference on human factors in computing systems. pp. 1–12 (2020) 5, 10
23. GmbH, P.: 10 types of powerpoint slides (2023), https://www.presentationload.com/blog/10-types-of-powerpoint-slides/ 4
24. Han, Y., Zhang, C., Chen, X., Yang, X., Wang, Z., Yu, G., Fu, B., Zhang, H.: Chartllama: A multimodal llm for chart understanding and generation (2023) 3
25. Haurilet, M., Al-Halah, Z., Stiefelhagen, R.: Spase-multi-label page segmentation for presentation slides. In: 2019 IEEE Winter Conference on Applications of Computer Vision (WACV). pp. 726–734. IEEE (2019) 2, 4, 8
26. Haurilet, M., Roitberg, A., Martinez, M., Stiefelhagen, R.: Wise—slide segmentation in the wild. In: 2019 International Conference on Document Analysis and Recognition (ICDAR). pp. 343–348. IEEE (2019) 2, 8
27. Hsu, H.Y., He, X., Peng, Y., Kong, H., Zhang, Q.: Posterlayout: A new benchmark and approach for content-aware visual-textual presentation layout. In: Proceedings of the IEEE/CVF Conference on Computer Vision and Pattern Recognition. pp. 6018–6026 (2023) 4
28. Hu, H., Chan, K.C., Su, Y.C., Chen, W., Li, Y., Sohn, K., Zhao, Y., Ben, X., Gong, B., Cohen, W., et al.: Instruct-imagen: Image generation with multi-modal instruction. In: Proceedings of the IEEE/CVF Conference on Computer Vision and Pattern Recognition. pp. 4754–4763 (2024) 14
29. HuggingFaceM4: Websight (v0.1). https://huggingface.co/datasets/HuggingFaceM4/WebSight (2024), accessed: 2024-01-16 3

30. Huh, M., Peng, Y.H., Pavel, A.: Genassist: Making image generation accessible. In: Proceedings of the 36th Annual ACM Symposium on User Interface Software and Technology. pp. 1–17 (2023) 14
31. Huh, M., Yang, S., Peng, Y.H., Chen, X., Kim, Y.H., Pavel, A.: Avscript: Accessible video editing with audio-visual scripts. In: Proceedings of the 2023 CHI Conference on Human Factors in Computing Systems. pp. 1–17 (2023) 14
32. Jandaghi, P., Sheng, X., Bai, X., Pujara, J., Sidahmed, H.: Faithful persona-based conversational dataset generation with large language models (2023) 3
33. Jeronymo, V., Bonifacio, L., Abonizio, H., Fadaee, M., Lotufo, R., Zavrel, J., Nogueira, R.: Inpars-v2: Large language models as efficient dataset generators for information retrieval. arXiv preprint arXiv:2301.01820 (2023) 3
34. Jiang, Y., Schoop, E., Swearngin, A., Nichols, J.: Iluvui: Instruction-tuned language-vision modeling of uis from machine conversations (2023) 3
35. Jobin, K., Mishra, A., Jawahar, C.: Semantic labels-aware transformer model for searching over a large collection of lecture-slides. In: Proceedings of the IEEE/CVF Winter Conference on Applications of Computer Vision. pp. 6016–6025 (2024) 2, 8, 10, 11
36. Khan, A.R., Khan, S., Harouni, M., Abbasi, R., Iqbal, S., Mehmood, Z.: Brain tumor segmentation using k-means clustering and deep learning with synthetic data augmentation for classification. Microscopy Research and Technique **84**(7), 1389–1399 (2021) 3
37. Kim, J., Choi, Y., Kahng, M., Kim, J.: Fitvid: Responsive and flexible video content adaptation. In: Proceedings of the 2022 CHI Conference on Human Factors in Computing Systems. pp. 1–16 (2022) 1, 2, 4, 5, 7, 8, 11, 12
38. Krosnick, R., Lee, S.W., Laseck, W.S., Onev, S.: Expresso: Building responsive interfaces with keyframes. In: 2018 IEEE Symposium on Visual Languages and Human-Centric Computing (VL/HCC). pp. 39–47. IEEE (2018) 1
39. Kumar, R., Satyanarayan, A., Torres, C., Lim, M., Ahmad, S., Klemmer, S.R., Talton, J.O.: Webzeitgeist: design mining the web. In: Proceedings of the SIGCHI Conference on Human Factors in Computing Systems. pp. 3083–3092 (2013) 3
40. Lee, D.W., Ahuja, C., Liang, P.P., Natu, S., Morency, L.P.: Lecture presentations multimodal dataset: Towards understanding multimodality in educational videos. In: Proceedings of the IEEE/CVF International Conference on Computer Vision. pp. 20087–20098 (2023) 2, 5, 8, 11, 12
41. Lee, J., Peng, Y.H., Herskovitz, J., Guo, A.: Image explorer: Multi-layered touch exploration to make images accessible. In: Proceedings of the 23rd International ACM SIGACCESS Conference on Computers and Accessibility. pp. 1–4 (2021) 10
42. Lee, K., Joshi, M., Turc, I.R., Hu, H., Liu, F., Eisenschlos, J.M., Khandelwal, U., Shaw, P., Chang, M.W., Toutanova, K.: Pix2struct: Screenshot parsing as pretraining for visual language understanding. In: International Conference on Machine Learning. pp. 18893–18912. PMLR (2023) 11
43. Leiva, L.A., Hota, A., Oulasvirta, A.: Enrico: A dataset for topic modeling of mobile ui designs. In: 22nd International Conference on Human-Computer Interaction with Mobile Devices and Services. pp. 1–4 (2020) 3, 4, 5, 11, 12
44. Li, J., Li, D., Savarese, S., Hoi, S.: Blip-2: Bootstrapping language-image pretraining with frozen image encoders and large language models. arXiv preprint arXiv:2301.12597 (2023) 1
45. Liang, P.P., Lyu, Y., Fan, X., Wu, Z., Cheng, Y., Wu, J., Chen, L., Wu, P., Lee, M.A., Zhu, Y., et al.: Multibench: Multiscale benchmarks for multimodal representation learning. Advances in neural information processing systems **2021**(DB1), 1 (2021) 13

46. Lin, M., Chau, M., Cao, J., Nunamaker Jr, J.F.: Automated video segmentation for lecture videos: A linguistics-based approach. International Journal of Technology and Human Interaction (IJTHI) **1**(2), 27–45 (2005) 11
47. Liu, H., Li, C., Wu, Q., Lee, Y.J.: Visual instruction tuning. In: NeurIPS (2023) 2, 11
48. Microsoft: The 10-20-30 rule of powerpoint (2023), `https://www.microsoft.com/en-us/microsoft-365-life-hacks/presentations/10-20-30-rule-of-powerpoint` 4
49. Min, S., Lyu, X., Holtzman, A., Artetxe, M., Lewis, M., Hajishirzi, H., Zettlemoyer, L.: Rethinking the role of demonstrations: What makes in-context learning work? arXiv preprint arXiv:2202.12837 (2022) 5
50. Moran, K., Bernal-Cárdenas, C., Curcio, M., Bonett, R., Poshyvanyk, D.: Machine learning-based prototyping of graphical user interfaces for mobile apps. IEEE Transactions on Software Engineering **46**(2), 196–221 (2020). `https://doi.org/10.1109/TSE.2018.2844788` 3
51. Nguyen, V.Q., Suganuma, M., Okatani, T.: Grit: Faster and better image captioning transformer using dual visual features. In: Computer Vision–ECCV 2022: 17th European Conference, Tel Aviv, Israel, October 23–27, 2022, Proceedings, Part XXXVI. pp. 167–184. Springer (2022) 1
52. Peng, Y.H., Bigham, J.P., Pavel, A.: Slidecho: Flexible non-visual exploration of presentation videos. In: Proceedings of the 23rd International ACM SIGACCESS Conference on Computers and Accessibility. pp. 1–12 (2021) 2, 14
53. Peng, Y.H., Chi, P., Kannan, A., Morris, M.R., Essa, I.: Slide gestalt: Automatic structure extraction in slide decks for non-visual access. In: Proceedings of the 2023 CHI Conference on Human Factors in Computing Systems. pp. 1–14 (2023) 2, 11
54. Peng, Y.H., Hsi, M.W., Taele, P., Lin, T.Y., Lai, P.E., Hsu, L., Chen, T.c., Wu, T.Y., Chen, Y.A., Tang, H.H., et al.: Speechbubbles: Enhancing captioning experiences for deaf and hard-of-hearing people in group conversations. In: Proceedings of the 2018 CHI Conference on Human Factors in Computing Systems. pp. 1–10 (2018) 1
55. Peng, Y.H., Jang, J., Bigham, J.P., Pavel, A.: Say it all: Feedback for improving non-visual presentation accessibility. In: Proceedings of the 2021 CHI Conference on Human Factors in Computing Systems. pp. 1–12 (2021) 2, 14
56. Peng, Y.H., Lin, M.T., Chen, Y., Chen, T., Ku, P.S., Taele, P., Lim, C.G., Chen, M.Y.: Personaltouch: Improving touchscreen usability by personalizing accessibility settings based on individual user's touchscreen interaction. In: Proceedings of the 2019 CHI Conference on Human Factors in Computing Systems. pp. 1–11 (2019) 1
57. Peng, Y.H., Wu, J., Bigham, J., Pavel, A.: Diffscriber: Describing visual design changes to support mixed-ability collaborative presentation authoring. In: Proceedings of the 35th Annual ACM Symposium on User Interface Software and Technology. pp. 1–13 (2022) 2, 5, 10, 14
58. Rodriguez, J.A., Agarwal, S., Laradji, I.H., Rodriguez, P., Vazquez, D., Pal, C., Pedersoli, M.: Starvector: Generating scalable vector graphics code from images (2023) 3
59. Ros, G., Sellart, L., Materzynska, J., Vazquez, D., Lopez, A.M.: The synthia dataset: A large collection of synthetic images for semantic segmentation of urban scenes. In: Proceedings of the IEEE conference on computer vision and pattern recognition. pp. 3234–3243 (2016) 3

60. Schoop, E., Zhou, X., Li, G., Chen, Z., Hartmann, B., Li, Y.: Predicting and explaining mobile ui tappability with vision modeling and saliency analysis. In: Proceedings of the 2022 CHI Conference on Human Factors in Computing Systems. CHI '22, Association for Computing Machinery, New York, NY, USA (2022). https://doi.org/10.1145/3491102.3517497, https://doi.org/10.1145/3491102.3517497 3
61. Sharma, P., Shaham, T.R., Baradad, M., Fu, S., Rodriguez-Munoz, A., Duggal, S., Isola, P., Torralba, A.: A vision check-up for language models. In: Proceedings of the IEEE/CVF Conference on Computer Vision and Pattern Recognition. pp. 14410–14419 (2024) 3
62. Strizic, M.: App Screens Design (2024), https://decode.agency/article/app-screens-design/ 4
63. Support, M.: What is a slide layout? (2024), https://support.microsoft.com/en-us/office/what-is-a-slide-layout-99da5716-92ee-4b6a-a0b5-beea45150f3a 4
64. Swearngin, A., Li, Y.: Modeling mobile interface tappability using crowdsourcing and deep learning. In: Proceedings of the 2019 CHI Conference on Human Factors in Computing Systems. p. 1–11. CHI '19, Association for Computing Machinery, New York, NY, USA (2019). https://doi.org/10.1145/3290605.3300305, https://doi.org/10.1145/3290605.3300305 3
65. Tan, M., Pang, R., Le, Q.V.: Efficientdet: Scalable and efficient object detection. In: Proceedings of the IEEE/CVF conference on computer vision and pattern recognition. pp. 10781–10790 (2020) 8
66. Tan, Z., Beigi, A., Wang, S., Guo, R., Bhattacharjee, A., Jiang, B., Karami, M., Li, J., Cheng, L., Liu, H.: Large language models for data annotation: A survey (2024) 4, 14
67. Tanaka, R., Nishida, K., Nishida, K., Hasegawa, T., Saito, I., Saito, K.: Slidevqa: A dataset for document visual question answering on multiple images. arXiv preprint arXiv:2301.04883 (2023) 2
68. Thirunavukarasu, A.J., Ting, D.S.J., Elangovan, K., Gutierrez, L., Tan, T.F., Ting, D.S.W.: Large language models in medicine. Nature medicine **29**(8), 1930–1940 (2023) 3
69. Tian, Y., Fan, L., Isola, P., Chang, H., Krishnan, D.: Stablerep: Synthetic images from text-to-image models make strong visual representation learners (2023) 3
70. Tian, Y., Cui, W., Deng, D., Yi, X., Yang, Y., Zhang, H., Wu, Y.: Chartgpt: Leveraging llms to generate charts from abstract natural language. IEEE Transactions on Visualization and Computer Graphics (2024) 3
71. Tseng, T., Cheng, R., Nichols, J.: Keyframer: Empowering animation design using large language models. arXiv preprint arXiv:2402.06071 (2024) 3
72. Vinyals, O., Toshev, A., Bengio, S., Erhan, D.: Show and tell: A neural image caption generator. In: Proceedings of the IEEE conference on computer vision and pattern recognition. pp. 3156–3164 (2015) 1
73. Viswanathan, V., Zhao, C., Bertsch, A., Wu, T., Neubig, G.: Prompt2model: Generating deployable models from natural language instructions (2023) 3, 5
74. Wang, B., Li, G., Zhou, X., Chen, Z., Grossman, T., Li, Y.: Screen2words: Automatic mobile ui summarization with multimodal learning. In: The 34th Annual ACM Symposium on User Interface Software and Technology. pp. 498–510 (2021) 3, 10, 11
75. Wang, R., Zhou, W., Sachan, M.: Let's synthesize step by step: Iterative dataset synthesis with large language models by extrapolating errors from small models (2023) 3

76. Westerlund, M.: The emergence of deepfake technology: A review. Technology innovation management review **9**(11) (2019) 14
77. Wu, J., Barik, T., Zhang, X., Lea, C., Nichols, J., Bigham, J.P.: Reflow: Automatically improving touch interactions in mobile applications through pixel-based refinements. arXiv preprint arXiv:2207.07712 (2022) 1
78. Wu, J., Krosnick, R., Schoop, E., Swearngin, A., Bigham, J.P., Nichols, J.: Never-ending learning of user interfaces. In: Proceedings of the 36th Annual ACM Symposium on User Interface Software and Technology. pp. 1–13 (2023) 2, 3
79. Wu, J., Peng, Y.H., Li, A., Swearngin, A., Bigham, J.P., Nichols, J.: Uiclip: A data-driven model for assessing user interface design. arXiv preprint arXiv:2404.12500 (2024) 14
80. Wu, J., Schoop, E., Leung, A., Barik, T., Bigham, J.P., Nichols, J.: Uicoder: Finetuning large language models to generate user interface code through automated feedback. arXiv preprint arXiv:2406.07739 (2024) 7, 14
81. Wu, J., Wang, S., Shen, S., Peng, Y.H., Nichols, J., Bigham, J.P.: Webui: A dataset for enhancing visual ui understanding with web semantics. In: Proceedings of the 2023 CHI Conference on Human Factors in Computing Systems. pp. 1–14 (2023) 2, 9, 13
82. Wu, J., Zhang, X., Nichols, J., Bigham, J.P.: Screen parsing: Towards reverse engineering of ui models from screenshots. In: The 34th Annual ACM Symposium on User Interface Software and Technology. p. 470–483. UIST '21, Association for Computing Machinery, New York, NY, USA (2021). `https://doi.org/10.1145/3472749.3474763`, `https://doi.org/10.1145/3472749.3474763` 3
83. Wu, J., Tenenbaum, J.B., Kohli, P.: Neural scene de-rendering. In: Proceedings of the IEEE Conference on Computer Vision and Pattern Recognition. pp. 699–707 (2017) 3
84. Xu, K., Ba, J., Kiros, R., Cho, K., Courville, A., Salakhudinov, R., Zemel, R., Bengio, Y.: Show, attend and tell: Neural image caption generation with visual attention. In: International conference on machine learning. pp. 2048–2057. PMLR (2015) 1
85. Zhang, T., Kishore, V., Wu, F., Weinberger, K.Q., Artzi, Y.: Bertscore: Evaluating text generation with bert. arXiv preprint arXiv:1904.09675 (2019) 5
86. Zhang, X., de Greef, L., Swearngin, A., White, S., Murray, K., Yu, L., Shan, Q., Nichols, J., Wu, J., Fleizach, C., et al.: Screen recognition: Creating accessibility metadata for mobile applications from pixels. In: Proceedings of the 2021 CHI Conference on Human Factors in Computing Systems. pp. 1–15 (2021) 3, 8
87. Zhou, S., Xu, F.F., Zhu, H., Zhou, X., Lo, R., Sridhar, A., Cheng, X., Bisk, Y., Fried, D., Alon, U., et al.: Webarena: A realistic web environment for building autonomous agents. arXiv preprint arXiv:2307.13854 (2023) 3
88. Zhou, X., Koltun, V., Krähenbühl, P.: Probabilistic two-stage detection. arXiv preprint arXiv:2103.07461 (2021) 8
89. Zhu, X., Su, W., Lu, L., Li, B., Wang, X., Dai, J.: Deformable detr: Deformable transformers for end-to-end object detection. arXiv preprint arXiv:2010.04159 (2020) 8